# Efficient Hierarchical Markov Random Fields for Object Detection on a Mobile Robot

Colin Lea and Jason J. Corso

*Abstract*— Object detection and classification using video is necessary for intelligent planning and navigation on a mobile robot. However, current methods can be too slow or not sufficient for distinguishing multiple classes. Techniques that rely on binary (foreground/background) labels incorrectly identify areas with multiple overlapping objects as single segment. We propose two Hierarchical Markov Random Field models in efforts to distinguish connected objects using tiered, binary label sets. Near-realtime performance has been achieved using efficient optimization methods which runs up to 11 frames per second on a dual core 2.2 Ghz processor. Evaluation of both models is done using footage taken from a robot obstacle course at the 2010 Intelligent Ground Vehicle Competition.

## I. INTRODUCTION

Accurate identification of all objects in a scene is a difficult problem in mobile robotics. The sensors used for object detection on a small vehicle are often limited to a camera and planar laser measurement unit due to cost and space restrictions. While laser units generally have high accuracy, the data is restricted to a planar slice in front of the robot. There are many instances where objects in the scene appear out of the range of the sensor because they are above or below the height of the laser beam. Additionally, objects may have important texture information that is necessary for classification. These conditions make a strong case for the use of vision-based methods of detection and identification. Our work leverages prior research in the areas of segmentation and classification using probabilistic models and previous work in vehicle lane detection for extracting the foreground-background in an outdoor environment.

Our primary contribution lies in detection of objects from our environment. In our previous experiments the use of a single-layer Markov Random Field for detecting objects from an image was inadequate for differentiating multiple objects that were overlapping. We propose two methods using Hierarchical Markov Random Fields to segment multiple sets of objects from an image. We use two-tiers of binary labels to prevent overlapping objects from being segmented together. While the on-board processing power and a necessity for near-realtime performance limit the use of many traditional Bayesian methods, we show that using Iterated Conditional Modes for optimization provides sufficiently fast performance.

The proposed detection methods are run on video that has been preprocessed using a foreground/background seg-

C. Lea is with Department of Mechanical and Aerospace Engineering, SUNY at Buffalo, Buffalo, NY 14260 `colinlea@buffalo.edu`

J. Corso is with the Department of Computer Science and Engineering, SUNY at Buffalo, Buffalo, NY 14260 `jcorso@buffalo.edu`

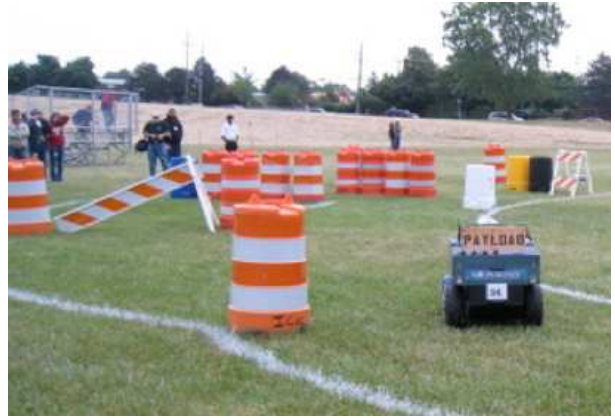

Fig. 1. Our vehicle navigating a robot obstacle course at the 2010 Intelligent Ground Vehicle Competition

mentation technique to get a binary image mask. Connected components are extracted and then classified with a decision tree-based technique that uses the conditional probability for each of the classes.

The overall goal is to be able to detect and classify features using a camera to aid in autonomous navigation of a robot obstacle course at the Intelligent Ground Vehicle Competition. By incorporating semantic knowledge about our surroundings, enhanced path planning strategies can be incorporated to better traverse the environment. In our outdoor course the robot must stay between white spray-painted lines while traveling to specified GPS locations. Figure 1 shows our robot navigating through the course at the 2010 Competition. Objects including traffic cones, multi-colored barrels, ramps, and other features are placed around the course. While we use a laser rangefinder for detecting some of the obstacles, it is unable to find parts of objects that are off of the ground such as a tilted sawhorse, seen on the left in figure 1.

In section 2 we look at prior work in the areas of lane detection and object segmentation. In section 3 we discuss preprocessing and formulate two models for a Hierarchical Markov Random Field and appropriate optimization and parameter estimation techniques. Section 4 compares the two proposed methods and Section 5 develops a high-level classifier for our experiment. Section 6 includes a discussion of our results.

## II. RELATED WORK

Markov Random Fields (MRF) are common in areas such as image restoration for their ability to estimate an original,

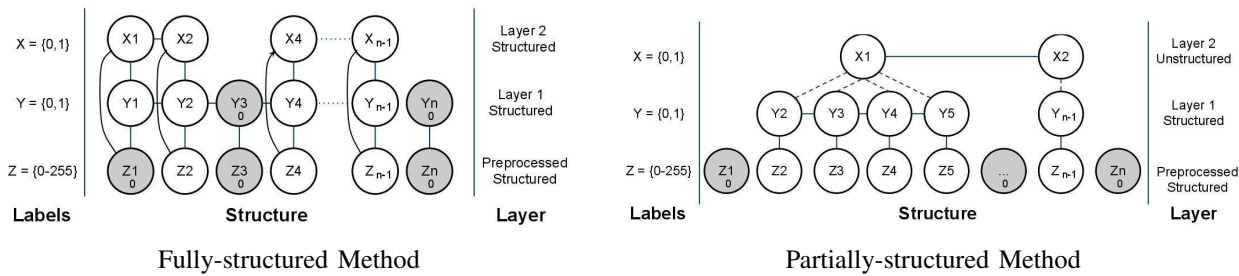

Fig. 2. Two Hierarchical Markov Random Field models are proposed for object detection on a mobile robot. In each, one layer is used to eliminate noise in the image and the other is used to aid inference by applying binary class labels. From bottom up, the fully-structured model on the left denoises and then classifies using structured layers. The partially-structured model on the right generates a binary object class and then uses an unstructured graph to denoise the class labels.

ideal image from an image with noise or errors [1]. They have also been used for a variety of additional applications such as object segmentation [2]. The largest problem with these is that most optimization techniques are too time intensive. For example graph cuts perform well in stereo reconstruction but ultimately are too computationally expensive for real-time application [3]. Additionally, because we are using binary labels we can not differentiate between multiple overlapping objects. Work using Hierarchical MRFs, such as [4], [5], [6], [7], has been proposed which can speed up the optimization process and increase robustness. By using multiple tiers of binary MRFs we can differentiate between multiple classes. Kanade *et al.* [4] present a hierarchy of MRFs that have successfully been used to infer the depth-relation between objects in a single image. Corso *et. al* have presented a two stage model for inferring lumbar disks in MRI images for a medical application [5]. In this paper we develop a different kind of hierarchy useful for efficient multiclass segmentation.

Due in part to the DARPA Grand and Urban challenges, significant research has been done recently in the area of autonomous road vehicles. Kastrinaki *et al.* summarizes a large body of work related to road detection using 2D and 3D techniques [8]. Template Matching has been shown to work moderately well using a monocular camera but comes at a significant computational cost and only marginal improvement over basic edge detection methods [9], [10]. In several studies, stereo vision has been used to combine the geometry of the road with traditional image processing methods to get more accurate results. Additional work demonstrates how texture analysis can be used to find the lanes on a road [11], [12].

While our work generalizes beyond the application for the Intelligent Ground Vehicle Competition, it is worth noting techniques other teams have taken in recent years. Traditionally at competition, groups have only focussed on detection of lane markers. The most common technique is to use an edge detector and to run a Hough transform to parameterize a curve. University of Detroit Mercy's and Princeton's teams have shown innovative solutions to the vision problem in recent years. The process that Detroit Mercy uses is as follows. Binary thresholds are used to estimate the location of the lanes and then a "Quad-Hough" transform parameterizes a line in each quadrant of the image [13]. They then use a Kalman Filter to smooth changes in the line over time. For lane detection, Princeton thresholds based on white values in the HSL colorspace and uses Random Sample Consensus (RANSAC) to fit parabolic lines. A stereo camera is used to detect objects in the scene and to identify a ground plane for the image data [14]. None of the teams investigated relied on more complicated multiclass techniques.

### III. LOW-LEVEL PROCESSING

Our preliminary goal is to isolate the lanes and features from the foreground in our images. In this section we discuss preprocessing methods and develop our primary segmentation techniques. We present two probabilistic approaches to detection using two-layered Markov Random Fields. Methods for optimization and parameter estimation are then discussed.

#### A. Preprocessing

Initial processing of the input image plays in important role in forming our model. We want to differentiate foreground-background from the camera's color image for use in feature detection. The saturation component of the Hue-Saturation-Luminance (HSL) colorspace was chosen as a basis for its superior ability to differentiate the lanes and objects from the background. However, there is a problem where shadows sometimes are too close in color to the features and trigger false positives. By combining information from both the saturation and luminance channels of HSL we are able to define a better initial image. A morphological opening filter is then applied to eliminate noisy pixels.

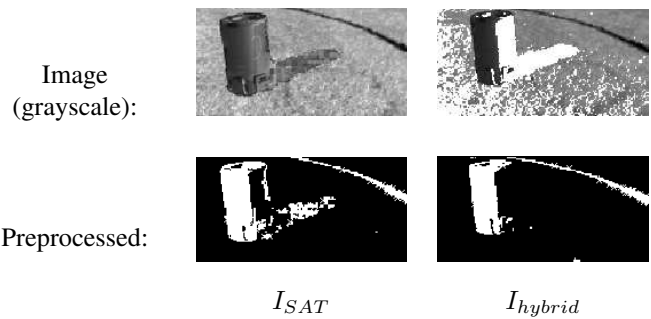

Fig. 3. Preprocessing is performed to eliminate the grassy background from the lanes and obstacles on the course.

$$I_{hybrid} = max(I_{SAT}, (I_{LUM} < \alpha)) \quad (1)$$

Figure 3 shows the saturation and hybrid channels and their respective preprocessed binary images. Both channels are thresholded at separate values, $\alpha_{\{S,L\}}$, which are based on the peaks in the histograms of each channel.

*B. Hierarchical Markov Random Field Framework (HMRF)*

Our initial work using a single-layer Markov Random Field for object detection on a robot showed that large amounts of noise resulted in many misclassifications. This is a problem because in our application the white lane markers are often faded or muddy and have the same texture as grass. Furthermore, using binary labels prevents the detection of multiple overlapping objects. Classification thus becomes a problem because features from two or more objects may be combined into one segment. Through the use of Hierarchical Markov Random Fields we are able to implement a more accurate and robust system for segmentation and classification.

Each of our models, shown in figure 2, has two steps: denoising and inference. What differentiates them is the order they perform these steps and the structure of their lattices. The fully structured method uses a stack of two traditional 4-connected single-layer MRFs. The partially structured HMRF has a structured MRF on the layer connected to the observed image data and an unstructured MRF on top joining the nearest connected neighbors. This is described in greater detail below.

The distinction from other hierarchical implementations such as [4] and [5] is in the definition of each layer. Kanade *et al.* use a pyramid of connected MRFs to propagate depth cues over the image. By using coarse-layers they are able to propagate information in fewer iterations. Corso *et al.* marginalize between the layers in the hierarchy in efforts to infer a label.

**Method I: Fully Structured**: In this method, Layer 1 denoises the image by operating on a binary representation of the preprocessed image. Errors resulting from the preprocessing stage are resolved based on interactions between each site's local neighborhood. This step is iterated until the nodes converge on a final result. The second layer is a function of the first layer's binary label and the corresponding grayscale values from the original image. A second binary label is generated on top of the foreground pixels to differentiate between objects. Thus, objects that appear to be touching in the image, which previously were labeled as single segments, can now be split up into multiple objects for classification.

**Method II: Partially Structured**: The second method generates a label using a structured MRF and then denoises the result with an unstructured graph. The first layer uses the grayscale values from foreground nodes in the preprocessed image and generates a binary label used to distinguish between different objects. The connected components for each label are extracted and input into an unstructured graph based on it's k nearest neighbors. The second layer reduces over-segmentation problems caused by noise in the underlying image.

*C. Markov Random Field Model*

Formulation, optimization and parameter estimation for each structured layer is done independently and in the same manner. Thus, we mathematically describe our model in terms of single-layer MRFs. Note that there are two differences for the unstructured layer in the second method. Instead of using an $n$ x $m$ lattice we simply join the nodes using the k nearest neighbors. Additionally, the unstructured MRF's parameters of are hand calibrated and not done using the coding method. For a more fundamental explanation of MRFs see [1].

A single-layer MRF is formed by a lattice containing an observation and a hidden layer. The fundamental principle assumes that each hidden value is dependent only on its corresponding pixel site and its hidden neighbors. In our case we want to find an ideal segmentation that only detects lanes and objects in the scene. We formulate an image lattice $\Lambda = \{s = (x, y) : 0 \leq x < n, 0 \leq y < m\}$ with binary labels $\mathcal{L} = \{-1, +1\}$ where -1 is background and +1 is foreground. The general form using a Potts model is given by:

$$P(f_i|f_{\mathcal{N}_i}) = \frac{\exp\left[-V_1(f_i) - \beta \sum_{i' \in \mathcal{N}} V_2(f_i, f_{i'})\right]}{\sum_{f_i \in \mathcal{L}} \exp\left[-V_1(f_i) - \beta \sum_{i' \in \mathcal{N}} V_2(f_i, f_{i'})\right]} \quad (2)$$

$V_1$ and $V_2$ are energy functions corresponding to the likelihood and smoothness prior respectively and $f$ is the hidden value at a site and d is the observed value. We evaluate both functions as variants of

$$V = \left(\frac{f_1 - f_2}{2}\right)^2. \quad (3)$$

The above equation gives a positive energy if the two values are different and zero if the values are the same. Note that for the first layer of Method II we use the normalized error from the 0-255 grayscale number but still evaluate to a binary label. The resulting form of our MRF is

$$P(\lambda_i|f_{\mathcal{N}_i}) = \frac{1}{Z} \exp\left[-(\frac{\lambda_i - d_i}{2})^2 - \frac{\beta}{|\mathcal{N}|} \sum_{i' \in \mathcal{N}} (\frac{\lambda_i - f_{i'}}{2})^2\right] \quad (4)$$

where Z is defined as the partition function and $\lambda_i$ is the label at a site. To more easily evaluate the system for different levels of connectivity we normalize using the number of connected neighbors, $|\mathcal{N}|$.

The likelihood function in the unstructured MRF of the second method is formulated in the same way as the structured MRFs, but the prior function is not. The prior is a function of the binary label, distance to the nearest neighbors, and pixel count at the neighbors. We are going to minimize this function, so the closer the neighbor is to a particular site the lower the value. Similarly, the the ratio of pixel counts is incorporated such that a larger neighbor decreases the value of the function. The rational is that a larger segment that is close in proximity is likely to be part of the current object. The energy function for the prior component is given by

$$V_i = \sum_{i' \in \mathcal{N}} \left(\frac{f_i - f_{i'}}{2}\right)^2 \cdot \frac{S_i}{S_{i'}} \cdot \frac{D_{i,i'}}{D_{max}} \quad (5)$$

where $S_i$ is the pixel count, $D_{i,i'}$ is the distance to a neighbor, and $D_{max}$ is the maximum distance from the site.

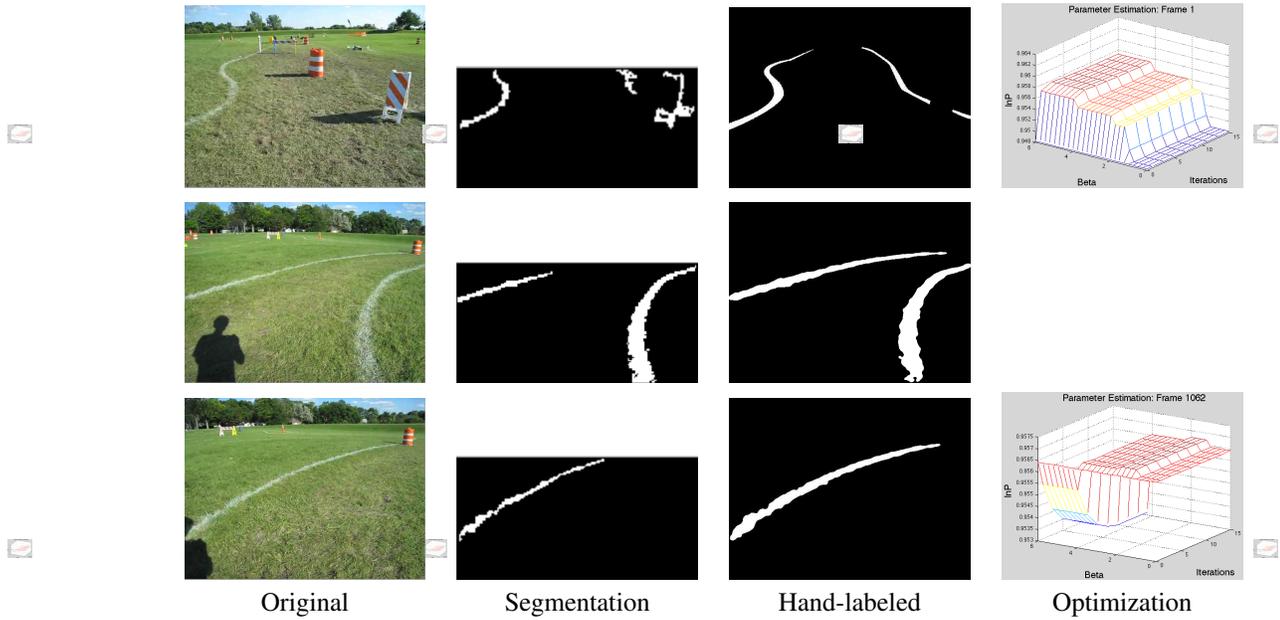

|  Original  |  Segmentation  |  Hand-labeled  |  Optimization  |

Fig. 4. The parameter estimation procedure is shown for Method I (fully-structured), Layer 1. The $\beta$ value and number of iterations of the MRF are compared in the graph on the right. For most images the image becomes stationary at two iterations with a $\beta$ of 1.8.

### D. Optimization

Traditionally, methods such as simulated annealing and belief propagation are used for accurate optimization of a Markov Random Field. However, these methods are time intensive and are not feasible for real-time applications on a mobile robot. Instead, Iterated Conditional Modes (ICM) is used for its fast convergence rate. While this technique is greedy and, in general, performs poorly for reconstructing large erroneous regions [2], it is sufficient for our situation. Mislabeled points in our experiments can appear as shot noise and thus can be eliminated without relying on more sophisticated algorithms. ICM deterministically calculates which of a set of labels has the lowest energy at each lattice site:

$$f_i = \arg\min_{\mathcal{L}_i} \left[\left(\frac{\lambda_i - d_i}{2}\right)^2 + \frac{\beta}{|\mathcal{N}|} \sum_{i' \in \mathcal{N}} \left(\frac{\lambda_i - f_{i'}}{2}\right)^2\right] \quad (6)$$

For our binary case both +1 and -1 are tested. This method should be iterated until the there are no changes to the image. As shown in figure 4 there is little change in the results after 2 iterations; thus, for computational reasons we preset the number of iterations to 2.

### E. Parameter Estimation

The coding method [15] is employed to estimate the parameter $\beta$ (the neighborhood component in the MRF). This method simplifies the estimation process by assuming neighborhood independence. The negative log likelihood of our model is used to compare our output with an accurately labeled image. Because of the neighborhood independence assumption this likelihood simplifies to:

$$-ln P^{(k)}(f|\theta) = \left(\frac{f_i - d_i}{2}\right)^2. \quad (7)$$

The value at each site is averaged over the lattice to find the likelihood of each test parameter where K is the total number of sites and $P^{(k)}$ is the probability at an individual site that is part of the lattice.

$$\beta^* = \frac{1}{K} \sum_k \arg\max_\beta P^{(k)}(f|\theta) \quad (8)$$

The comparison in figure 4 refers to parameter estimation for the first layer in the fully-structured method and shows three sample images and their corresponding negative log-likelihood probabilities. We compare this with with the number of iterations of Iterated Conditional Modes. It is apparent that the optimal parameter values vary with the context of the scene. When there is a lot of noise in the image a high $\beta$ and high iteration number are best – increasing iterations causes erroneous points to disappear. However, other times a high iteration count with a high $\beta$ can cause too much erosion in the image. The tradeoff appears to be noise vs. coherent object structure. Over a series of test images we find a sufficient $\beta$ value of 1.8. The number of iterations is dependent on the pre-processed image. On most test images a lower value is more accurate than a higher value.

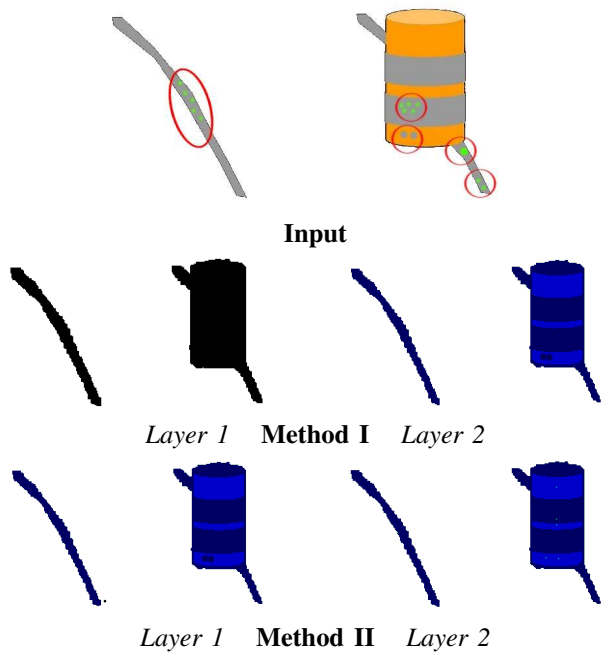

Fig. 5. This test image was generated to highlight the differences between our methods at each stage in the hierarchy. The first method denoises the foreground-background and then provides a binary label based on the grayscale values of the foreground. The second method provides a binary label based on the grayscale values and then denoises the connected components using an unstructured graph.

| Frame | $\beta$ |
|-------|---------|
| A | $\geq 4$ |
| B | 1 |
| C | 1.5 |

TABLE I. Method I, Layer 1 Parameters

## IV. EXPERIMENT: TEST IMAGE

In order to show the differences between the output of our two models, we created a test image, figure 5, to highlight how each stage functions. This image was generated to determine how effective each method is in eliminating noise and providing the most accurate segmentation. The test image includes three objects: a striped traffic barrel and two lanes. The right-most lane is running behind the barrel and would result in a single object using a single-layered binary segmentation. Noise was added to multiple areas in the image. The red circles in the "Input" are used to show the "erroneous" areas that simulate mud or other discolored areas in our image – the circles are not actually shown in the test images.

Results from the test are shown in figure 5. As expected the first layer of Method I (fully-structured) shows a mask outlining all of the objects in the image. Grass-colored noise that was added on top of the lines and barrel are correctly denoted as foreground. In the second layer the larger pieces of noise on the bottom of the barrel are mislabeled. This follows our intuition that using Iterated Conditional Modes will successfully get rid of shot noise but not larger regions. It is expected that these small bits would be marked as noise in our classification stage, but this does not result in an accurate segmentation.

As expected, the first layer of Method II (partially-structured) over-segments the image. Some of the noise is labeled as a separate texture than the surrounding nodes. In the second layer the noise and actual segments are merged into one piece. Based on this example it is apparent that the second method can in some cases provide superior results compared to Method I. Later we look at results of both methods on video.

## V. HIGH-LEVEL PROCESSING

Our end goal is to be able to classify objects in our environment, most importantly delineating the difference between lanes and objects. Using the connected components from our previous models we seek to identify objects from a list of 4 types: left lane, right lane, traffic fixtures, and ramps. We acknowledge that there may still be noise (either shot noise or large segments), so we must have a way of also classifying erroneous features. We present a decision tree that uses Bayesian methods to classify each segment.

### A. Decision Tree

Several classification techniques were investigated and ultimately we found that a simple Decision Tree is adequate for identifying our objects. There is a large difference between the characteristics of each object type: lanes are long and skinny and other objects are more rectangular. Using the empirical evidence shown in Figure 6 we generated a tree, Figure 7, that provides the best inference on the data. The feature used at each branch is based on the largest deviation between classes from our training set. The list of potential features includes pixel count, pixel to area ratio, length in each axis, and others discussed below.

At each branch we find the conditional probability of a segment being in the label set $\mathcal{L} \in$

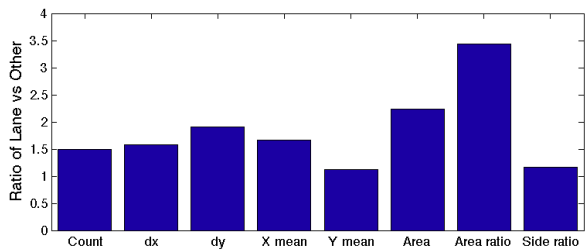

Fig. 6. The ratio of the values for Lanes and Objects is used to identify the feature that best differentiates each class. This shows that the area ratio is the best indicator of whether a segment is a lane or object and thus is used in the first branch of the decision tree.

$\{Object_1, Object_2, Error\}$. The farther away a feature is from the average for each given object, the lower the probability. The prior probability of each feature is learned and incorporated into our model to increase robustness. S is denoted as a segment, N is the feature value, and f is the particular feature.

$$L(S|\mathcal{L}_i) = \exp\left[-\frac{|(N_{f,\mathcal{L}_i} - N_{f,S})|}{N_{f,\mathcal{L}_i}}\right] \quad (9)$$

This is combined with the probability that any segment is of a particular label. The maximum probability is chosen as a segment's label.

$$S = \arg\max_{\mathcal{L}_i} \; L(S|\mathcal{L}) \cdot P(\mathcal{L}) \quad (10)$$

Segments were hand-labeled to evaluate the size and shape features above. We find that there is greatest deviation between the lanes and all other objects.

*B. Classification Results*

Video data of the entire practice course at the 2010 Intelligent Ground Vehicle Competition was recorded using a handheld camera. Using this video we have evaluated our ability to classify objects. Results from two sections of the course are shown in the following table. Data is evaluated every 50 frames in the sequence totaling 213 segments over 68 frames.

| Region: | "Good" | "Poor" | Overall |
|---|---|---|---|
| Segments: | 1-110 | 111-213 | 1-213 |
| Accuracy: | 93.0% | 45.8% | 70.0% |

TABLE II. Classification Accuracy

The labels generated from the decision tree are compared to hand labeled data to calculate the accuracy. The differentiation between "Good" and "Poor" largely has to do with the amount of noise in the image from the preprocessor.

## VI. DISCUSSION

Figure 8 depicts a set of samples from the video. Item A depicts a situation where the fully-structured method provides better results and B–E show areas where the unstructured method is superior. Item A is a near-perfect example of why Hierarchical MRFs are effective for segmenting multiple overlapping objects. Simply using foreground-background labels would identify the left lane and barrel as a single object. Note that the actual color of each segment does not matter – color only denotes separated regions.

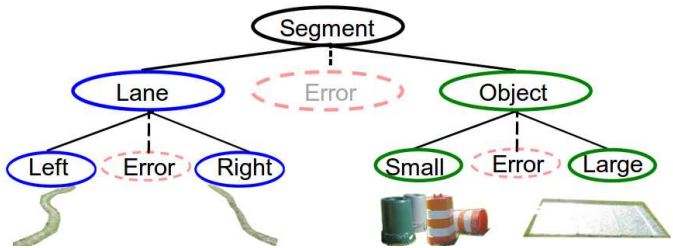

Fig. 7. A Decision Tree is used to identify each of our four classes. Error can be characterized in different ways based on our image and is included at each stage in our tree.

In item B, the first method over-segments the right lane marker. There are yellow spots surrounding the lane that are eliminated using the second model. Because the pixel size of the small yellow regions is small compared to the large segment, the unstructured method combines them with the lane. Similarly, in item C the base of the traffic fixture merges with the rest of the of the object. It is worth pointing out that part of the cone and white lane are detected a a single object. This has to do with the fact that both features are white in texture.

In item E the left lane is inaccurately labeled using the first method and the right lane is inaccurately labeled using the second method. In areas with clustered barrels, such as D, our method does a good job of separating different parts of the objects. While this may decrease the reliability of our classification technique, it is superior to our previous method using single binary labels which classified all of the barrels together as one unit.

As previously discussed, noise takes multiple forms. This is best shown in items A and E. In A, erroneous segments are relatively small and look like large shot noise. Conversely, in E the top portion of the image features a large error segment. This is why it is necessary to include the noise component at each stage of the decision tree.

Methods I and II run at about 11 and 6 frames per second respectively on quarter-scale VGA images using a dual core 2.2 GHz processor. The actual frame rate is dependent on the scene. For example, when calculating the nearest neighbors to form an unstructured graph in Method II, layer 2, having a large number of segments will result in longer computation time and thus slower frame rates. In general Method II stays between 3 and 9 frames per second. It has been coded using Python with the NumPy numerical library. It is possible to speed up the implementation by optimizing in C++ or incorporating a graphics card to offset computational cost.

One problem with our classification method is robustness through time. Especially when an object is nearing the edge of the image, an "Object" label is sometimes changed to "Lane." This is because of the changing feature vector – as an object is occluded its geometric shape becomes thinner and it starts looking more like a lane. By tracking the labels through time we can potentially avoid this problem and retain the original classification. Additionally, improvements such as a more robust method of foreground-background preprocessing may provide for better experiments with our segmentation methods.

Based on our results from both the generated image and video dataset, both methods show potential for future applications. In the experiment in section 3 we see that Method II provides a more proper segmentation, but in the video there are some situations with conflicting results. Considering the second method is more computationally expensive, whether or not it should be chosen over the first model may depend

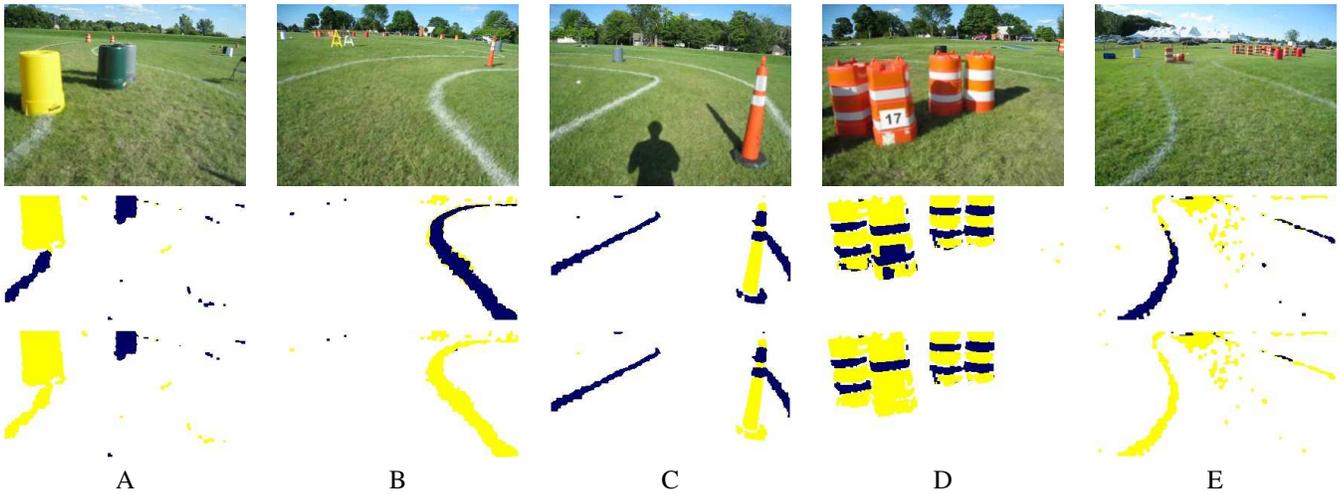

Fig. 8. (top) Original images (middle) Method I (bottom) Method II. Result are shown from video of the robot obstacle course at the Intelligent Ground Vehicle Competition. The yellow and dark blue colors denote different connected regions using our two proposed Hierarchical Markov Random Fields.

on the application.

VII. CONCLUSION

We have proposed two formulations of a Hierarchical Markov Random Field for segmenting objects in video and have evaluated them on footage from the Intelligent Ground Vehicle Competition. This work shows that Hierarchical MRFs can be implemented efficiently for real-time application. Results using a decision tree have proven successful in well defined areas on the obstacle course. Tests on our example image shows that the first method was only capable of eliminating shot noise but the second method was capable of denoising larger mislabeled segments. In general we find that Method II performs better but at some computational cost.